\documentclass{article}
\usepackage{framed,multirow}
\usepackage{amsmath}
\usepackage{amsfonts, amssymb,amsthm}

\usepackage{times}
\usepackage{graphicx} 
\usepackage{subfigure} 

\usepackage[numbers]{natbib}

\usepackage{algorithm}
\usepackage{algorithmic}

\usepackage{hyperref}

\usepackage{comment}
\usepackage{soul}

\usepackage{epstopdf}

\usepackage{multirow}

\usepackage[colorinlistoftodos, textwidth=40mm, shadow]{todonotes}

\newcommand{\beq}{\begin{equation}}
\newcommand{\eeq}{\end{equation}}

\newcommand{\PP}{\mathbb{P}}
\newcommand{\RR}{\mathbb{R}}
\newcommand{\EE}{\mathbb{E}}

\newcommand{\cc}{\mathcal{C}}

\newcommand{\pt}{\tau}

\newcommand{\tkernel}{K^h}

\theoremstyle{plain}
\newtheorem{theorem}{Theorem}

\theoremstyle{remark}

\theoremstyle{plain}

\theoremstyle{plain}

\theoremstyle{plain}
\newtheorem{proposition}[theorem]{Proposition}
\theoremstyle{definition}

\begin{document} 

\author{
Thomas Bonis and Steve Oudot\\
DataShape Team\\
Inria Saclay\\
}

\title{A Fuzzy Clustering Algorithm for the Mode-Seeking Framework}
\maketitle

\begin{abstract}
In this paper, we propose a new fuzzy clustering algorithm based on the mode-seeking framework. 
Given a dataset in $\RR^d$, we define 
regions of high density that we call cluster cores. We then consider a random walk on a neighborhood graph built on top of our data points 
which is designed to be attracted by high density regions. The strength of this attraction
is controlled by a temperature parameter $\beta > 0$. The membership of a point to a given cluster 
is then the probability for the random walk to hit the corresponding cluster core before any other. 
While many properties of random walks (such as hitting times, commute distances, etc\dots) have been shown to enventually encode purely local information when the number of data points grows, 
we show that the regularization introduced by the use of cluster cores solves this issue. 
Empirically, we show how the choice of $\beta$ influences the behavior of our algorithm: for small values 
of $\beta$ the result is close to hard mode-seeking whereas when $\beta$ is close to $1$ the result is similar to the output of a (fuzzy) spectral clustering. 
Finally, we demonstrate the scalability of our approach by providing the fuzzy clustering of a protein configuration dataset containing a million data points in $30$ dimensions. 
\end{abstract}

\section{Introduction}
The analysis of large and possibly high-dimensional datasets is
becoming ubiquitous in the sciences. The long-term objective is to
gain insight into the structure of measurement or simulation data, for
a better understanding of the underlying physical phenomena at
work. Clustering is one of the simplest ways of gaining such insight,
by finding a suitable decomposition of the data into clusters such
that data points within a same cluster share common (and, if possible,
exclusive) properties.

In this work, we are interested in the mode seeking approach to
clustering. This approach assumes the data points to be drawn from
some unknown probability distribution and defines the clusters as the
basins of attraction of the maxima of the density, requiring a
preliminary density estimation
phase~\citep{YCEJS,Tomato,MeanShift,Graphrandomwalk,comaniciu2002mean,ModeSeeking}. 
The theoretical analysis of this clustering framework has drawn increasing attention recently, see 
\cite{YCAS, YCriskbounds, YCdensity, YCMorseSmale, Arias-Castro}.
However, this (hard) clustering method provides a fairly limited knowledge on
the structure of the data: while the partition into clusters is well
understood, the interplay between clusters (respective locations,
proximity relations, interactions) remains unknown.  Identifying
interfaces between clusters is the first step towards a higher-level
understanding of the data, and it already plays a prominent role in
some applications such as the study of the conformations space of a
protein, where a fundamental question beyond the detection of
metastable states is to understand when and how the protein can switch
from one metastable state to another~\citep{cspd-lfdsmds-06}.  Hard
clustering can be used in this context, for instance by defining the
border between two clusters as the set of data points whose
neighborhood (in the ambient space or in some neighborhood graph)
intersects the two clusters, however this kind of information is by
nature unstable with respect to perturbations of the data.

fuzzy clustering appears as the appropriate tool to deal with interfaces
between clusters. Instead of assigning each data point to a single
cluster, it computes a degree of membership to each cluster for each
data point. The promise is that points close to the
interface between two clusters will have similar degrees of membership
to these clusters.  Thus,
fuzzy clustering uses a fuzzier notion of cluster membership in order
to gain stability on the locations of the clusters
boundaries.

\begin{figure}
\label{cluscore}
\centering
\includegraphics[width=0.7\linewidth]{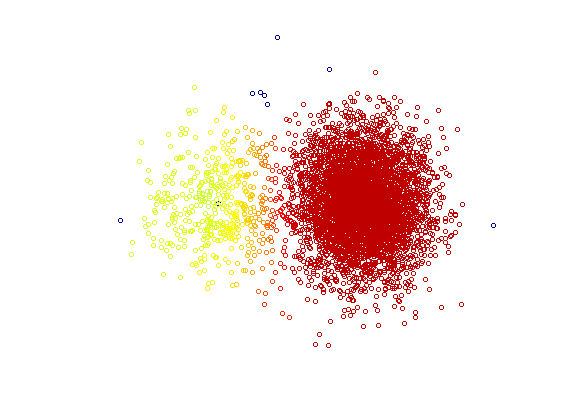}
\caption{Fuzzy clustering output for an unbalanced mixture of gaussian. Red color corresponds to the right cluster, blue to the left one. Finally,  
green points have similar membership to both clusters.}
\end{figure}

Consider a smooth density $f$ in $\RR^d$. Under the mode seeking paradigm, 
clusters correspond to the modes of $f$. More precisely, considering the gradient flow induced by f: 
\[
y'(t) = \nabla f (u(t))
\]
two points $x$ and $y$ are in the same cluster if the gradient flow
started at $x$ and the gradient flow started at $y$ have the same
limit which is a local maximum of $f$.  A natural way to turn
this approach into a fuzzy clustering algorithm is to follow a
perturbed gradient flow instead, such as the diffusion process
solution of 
\beq
\label{eq:diff-proc} dY_t = \frac{1}{\beta} \nabla(
\log f) dt + dB_t, 
\eeq 
where $B_t$ is a $d$-dimensional Brownian
motion and $\beta$ is a temperature parameter controlling the amount
of noise introduced in the gradient flow.  We use the gradient of the
logarithm of $f$ here as this quantity arises naturally in
practice. Indeed, since we only have access to a discretization of the
space through the sampled data points, we mimic this perturbed
gradient flow by a random walk on the data points.
\citet{Jordan} proved an isotropic random walk on a neighborhood
graph approximates the previous diffusion process for $\beta = 1$ while 
other values of $\beta$ are obtained by putting weights on the edges of the graph.
At this point, one could perform fuzzy clustering by considering the first local maximum of the
density encountered by the random walk, an approach wich has been proposed by~\citet{YCEJS}. 
However, as emphasized by~\citet{gettinglost}, the hitting time to a single point for a random walk on the graph
converges to
irrelevant quantities when the number of data points goes to infinity. We can thus expect the clustering to fail in that case. 
Indeed, if we apply this method to the fuzzy clustering of two
different Gaussian measures (see Figure~\ref{cluscore}). The obtained fuzzy
memberships are unsatisfying. In order to circumvent this issue, we assign a zone of high
density to each cluster, called \emph{cluster core} and computed
using the mode-seeking (hard) clustering algorithm
ToMATo~\citep{Tomato}. The fuzzy membership of a point to a given cluster is then given by 
the probability for the random walk started at this point to hit the corresponding cluster core first. 



\section{The Algorithm}
Our algorithm is a fuzzy generalization of the ToMATo algorithm which relies on the concept of $\emph{prominence}$. Let $G$ be a graph and $f$ be a real valued function on the vertices of this graph. 
For any $\alpha \in \RR$, let $F^\alpha = f^{-1}([\alpha, +\infty])$ be the $\alpha$-superlevel-set of $f$. A new connected component $C$
is born in $F^{\alpha}$ when $\alpha$ reaches a local maximum of $f$ on $G$ and we denote by $\alpha_{b,C}$ the corresponding value of $\alpha$. This component then dies at $\alpha = \alpha_{d,C} < \alpha_{b,C}$ when it gets connected, in $F^\alpha$, to another connected component $C'$ such that $\alpha_{b,C'} > \alpha_{b,C}$.
The prominence of $C$ (and by extension, of the corresponding local maximum of $f$) is then simply $\alpha_{b,C} - \alpha_{d,C}$.

The algorithm takes as input a finite set of points $\mathcal{X} =
\{X_1, \cdots, X_n\}$ together with pairwise distances $d(X_i,
X_j)$. In practice only the distances are used, so there is no need
for point coordinates.
Additionally, the algorithm takes in the following set of parameters:
\begin{itemize}
\item a density estimator $\hat{f}:\mathcal{X}\to\RR$,
\item a kernel $k$, for example the Gaussian kernel,
\item a window size $h>0$,
\item a prominence threshold $\pt>0$,
\item a temperature $\beta > 0$.
\end{itemize}
The first four parameters are in fact required by ToMATo for hard
mode-seeking, upon which our algorithm relies. The last parameter is
the one added in for fuzzy mode-seeking, as per
Equation~(\ref{eq:diff-proc}).


\noindent Given this input, our algorithm proceeds as follows:
\begin{enumerate}
\item It builds a weighted neighborhood graph~$G$  on top of the point
  cloud~$\mathcal{X}$, adding an edge with weight
\beq
\label{eq:poids}
w_{i,j} = \left(1 + \frac{1-\beta}{\beta} \hat{f}(X_j)\right)k\left(\frac{d(X_i, X_j)}{h}\right)
\eeq
between each pair of points $(X_i, X_j)$. Remark that it is possible to replace our kernel-based graph by a nearest neighbour graph. 
%
\item It computes the cluster cores by running ToMATo with input
 $\mathcal{X}$, $d$, $\log(\hat{f})$, $\pt$ and the unweighted neighborhood
 graph $\bar G$ obtained from~$G$ by removing the edges with weights
 lower than $0.5 \max(k)$.
The output of ToMATo is a set of~$K$
clusters~$C_1, \cdots, C_K$. Each cluster~$C_i$
corresponds to the basin of attraction of some peak of~$\log(\hat{f})$ of
prominence at least~$\pt$ within~$\bar G$. Up to a reordering of the
data points, we can assume this peak to be~$X_i$.  The $i$-th cluster
core~$\cc_i$ is then taken to be the {\em highest and most stable
  part} of~$C_i$, defined formally as the connected component
containing $X_i$ within the subgraph of~$\bar G$ spanned by those
vertices $X_j$ such that $\log( \hat{f}) (X_j) > \log( \hat{f}) (X_i) - \pt/2$.
%
\item
It computes the fuzzy-membership values $\mu_{1}, \dots, \mu_K$ by
solving the linear system $A^T \mu = \mu$, where the matrix $A$ is defined by:
\[
A_{kl} = \begin{cases}
\delta_{kl} & \mbox{if $X_k$ belongs to some cluster core}\\
\tkernel(X_k, X_l) & \mbox{otherwise},
\end{cases}
\]
where $\tkernel$ is the transition kernel of the random walk on the graph, i.e.
\beq
\label{eq:trans}
\tkernel(X_i,X_j) = \frac{w_{i,j}}{\sum_z w_{i,z}}.
\eeq
\end{enumerate}
The output of the algorithm is the set of fuzzy-membership values
$\mu_{1}, \dots, \mu_K$ computed at step~3.

\section{Parameters selection}
\label{sub:pract}

\subsection{Density estimator, window size, kernel and prominence threshold}
These $4$ parameters are tied to the classical hard mode-seeking
framework.  
The density estimator can be linked to the window size in practice, as
is done e.g. in Mean-Shift \citep{MeanShift} and its successors. For
instance, one can consider the kernel density estimator associated to the kernel $k$. 
This not only reduces the number of
parameters to tune in practice, but it also gives a way to select~$h$
using standard parameter selection techniques for density estimation, 
which is done for example in \cite{YCEJS}. Finally, the prominence threshold $\pt$ is used to
distinguish between relevant and irrelevant peaks in the discrete
setting.
It can be selected by running ToMATo twice: once to get the
distribution of prominences of the peaks of $\hat f$
within the neighborhood graph~$\bar G$, from which $\pt$ can be
inferred by looking for a gap in the distribution; then a second time,
using the chosen value of~$\pt$, to get the final hard
clustering. This procedure is detailed in~\citet{Tomato}.

\subsection{Temperature parameter}
This parameter is standard in fuzzy clustering. Outputs
corresponding to large values of $\beta$ will tend to have smooth
interfaces between clusters, while small values of $\beta$ will
encourage quick transitions from one cluster to another.  
$\beta$ can also be interpreted as a trade-off between the respective
influence of the metric and of the density in the diffusion process:
when $\beta$ is small, the output of our algorithm is mostly guided by
the density and therefore close to the output of mode seeking
algorithms; by contrast, when $\beta$ is large, the algorithm becomes
oblivious to the density. In practice, one may get insights into the choice of
$\beta$ by looking at the evolution of a certain measure of {\em
  fuzziness} of the output clustering across a range of values
of~$\beta$. We elaborate on this in Section~\ref{Experiments}.
 
\section{Convergence guarantees}
\label{sec:guarantees}

In this section we provide guarantees to our fuzzy clustering scheme by
exploiting the convergence of the random walk over the neighborhood
graph to a continuous diffusion process.  

As is usual in mode-seeking, we assume our input data points
$X_1,...,X_n$ to be i.i.d random variables drawn from some unknown
probability density~$f$ over~$\RR^d$. We also assume that the metric~$d$
that equips the data points is the Euclidean norm, and  that $f$ satisfies the
following technical conditions:
\begin{itemize}
\item $f$ is Lipschitz continuous over $\RR^d$ and $C^1$-continuous
  over the domain $\Omega = \{x \in \RR^d \mid f(x) > 0\}$,
\item $\lim\limits_{\|x\|_2 \to \infty} f(x) = 0$,
\item The SDE~\ref{eq:diff-proc} is well-posed. 
\end{itemize}
Standard sufficient conditions ensuring the well-posedness
(particularly the non-explosion) of the SDE~\ref{eq:diff-proc} can be found
in~\citet{Feller} or in~\citet{Strongsol}, for example one can assume $\nabla \log f$ to be Lipschitz continuous.

Our analysis connects random walks on graphs built on top of the input
point cloud~$\mathcal{X}$ using a density estimator~$\hat f$ to the
solution of Equation~\ref{eq:diff-proc}, for a fixed temperature parameter
$\beta > 0$. Specifically, let $M^{x,h}$ denote the Markov Chain whose
initial state is the closest neighbour of $x$ in the point
cloud~$\mathcal{X}$ (break ties arbitrarily), and whose transition
kernel $\tkernel$ is given by Equation~\ref{eq:trans}.  Following the
approach of~\citet{Jordan}, we show that, 
under suitable conditions on the estimator $\hat f$, this graph-based
random walk approximates the diffusion process in the continuous
domain in the following sense: there exists $s$ depending on $h$ such
that, as $n$ tends to infinity, with high probability, $M_{\lfloor
  t/s\rfloor}^{x,h}$ converges weakly to the solution
of Equation~(\ref{eq:diff-proc}).
From there, under standard conditions for mode estimation on the window size~$h$ and on
the density estimator~$\hat f$ (see \cite{YCMorseSmale, Arias-Castro}), we obtain the convergence of the
fuzzy-membership values~$\mu_{i}(x)$ computed by the algorithm to the
membership defined from the underlying continuous
diffusion process $\tilde{\mu}_i(x)$. 
Formally, letting $v_1, \dots, v_K$ be the
local maxima of $f$ of prominence higher than $\pt$, and $\tilde
\cc_1, \cdots, \tilde \cc_K$,  their associated cluster cores in the
continuous domain (i.e. $\tilde \cc_i$ is the connected component containing $v_i$ in $\{x \in \RR^d | \log f(x) \geq \log f(v_i) - \pt/2 \}$),
we define~$\tilde{\mu}_i(x)$ as the probability
for the diffusion process solution of~(\ref{eq:diff-proc}) to hit
$\tilde \cc_i$ before any other $\tilde \cc_j$.

\begin{theorem}
\label{fuzzycor}
Let $\beta > 0$ 
and assume $\|\nabla f\|$ is bounded from below on the boundary of the underlying cluster cores $\tilde{C}$.
Let $h : \mathbb{N} \rightarrow \RR^+$ be a decreasing window size such that $\lim\limits_{n \to \infty} h(n) = 0$ while $\lim\limits_{n \to \infty} \frac{h(n)^{d+2} n}{\log n} = \infty $. 
Suppose the density estimator $\hat{f}_n$ satisfies, for any compact set $U \subset \Omega$ and any $\epsilon > 0$,
\[ 
\lim\limits_{n \to \infty} \mathbb{P} (\sup_{x \in U} |\nabla f(x) - \nabla \hat{f}_n (x) | \geq h(n)^2 \epsilon) = 0.
\]
Then, for any compact set $U\subset\Omega$, 
any $\epsilon > 0$ and any $i$,
\[
\lim\limits_{n \to \infty} \mathbb{P}\left(\sup_{x\in U}|\mu_{i}(x) - \tilde{\mu}_i(x)| \geq \epsilon\right) = 0.
\]

\end{theorem}

\section{Experiments}
\label{Experiments}

We first illustrate the effect of the temperature parameter $\beta$ on
the clustering output using synthetic data. We then apply our method
on a couple UCI repository datasets and on simulated protein
conformations data.  In all our experiments we use a $k$-nearest
neighbor graph along with a distance to measure density estimator
\citep{distance2measure} computed using the $2k$ nearest-neighbors.

\subsection{Synthetic data}

\begin{figure}
  \centering

\subfigure[The data.]{
  \includegraphics[width=0.3\linewidth]{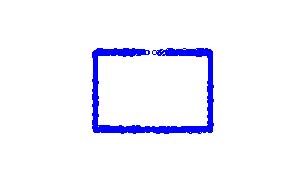}
  \label{Exp1data}
}
\subfigure[Output of ToMATo.]{
  \includegraphics[width=0.3\linewidth]{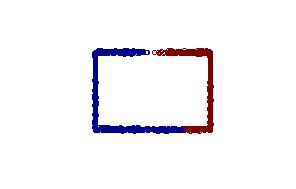}
  \label{Exp1hard}
}
\subfigure[Output for $\beta = 0.2$.]{
  \includegraphics[width=0.3\linewidth]{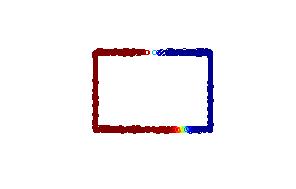}
  \label{Exp102}
}
\subfigure[Output for $\beta = 1$.]{
  \includegraphics[width=0.3\linewidth]{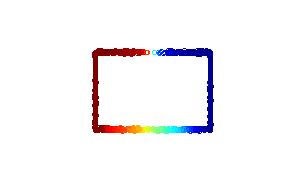}
  \label{Exp11}
}
\subfigure[Output for $\beta = 2$.]{
  \includegraphics[width=0.3\linewidth]{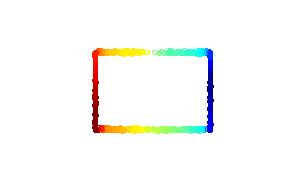}
  \label{Exp12}
}
\subfigure[Evolution of $H$ with respect to $\beta$.]{
  \includegraphics[width=0.3\linewidth]{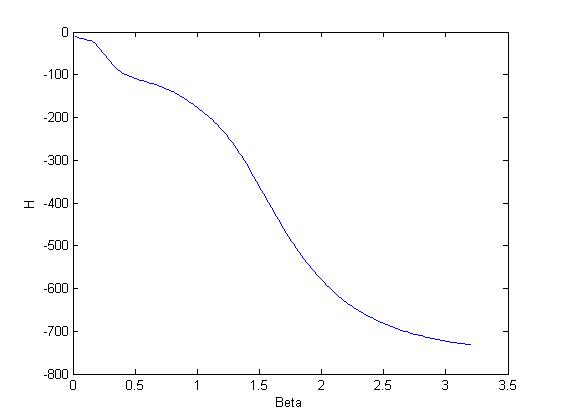}
  \label{fuzz}
}
\caption{Output of our algorithm on a simple dataset composed of two overlapping clusters. For fuzzy clustering green corresponds to an equal membership to both clusters.}
\end{figure}

The first dataset is presented in Figure~\ref{Exp1data} and is composed of two high-density clusters connected by two links. 
The bottom link is sampled from a uniform density while the top link is sampled from a density that has a gap inbetween the two clusters. 
Standard mode seeking
algorithms will have a hard time clustering the bottom link as a density estimation can create many ``noisy" local maxima: for instance, ToMATo
missclusters most of the bottom link (see Figure~\ref{Exp1hard}). We
display the results of our algorithm for three values of $\beta:$ $\beta = 0.2$
in Figure~\ref{Exp102}, $\beta = 1$ in Figure~\ref{Exp11} and $\beta = 2$ in
Figure~\ref{Exp12}. As we can see from the output of the algorithm, 
for small values of $\beta$, the amount of noise injected
in our trajectory is not large enough to compensate for the influence of the noise in the density estimation, so the result obtained is really close to hard
clustering. Large values of $\beta$ do not give enough
weight to the density function which leads to a smooth transition
between the two clusters on the top link.  Intermediate values of
$\beta$ seem to give more satisfying results. 
In order to gain intuition regarding which value of $\beta$ one should use, it is possible to look at the evolution of a fuzziness value for the clustering. 
For example, one can consider a notion of clustering entropy:
\beq
\label{eq:entropy}
H = \sum_i \sum_j \mu_j(X_i) \log(\mu_j(X_i)),
\eeq
which gets lower when the fuzziness of the clustering increases. As we can see in Figure~\ref{fuzz}, the evolution of $H$ with respect to $\beta$ presents three distincts 
plateaus corresponding to the three behaviour highlighted earlier. 

\begin{figure}
\subfigure[Output for $\beta = 1$.]{
  \includegraphics[width=0.3\linewidth]{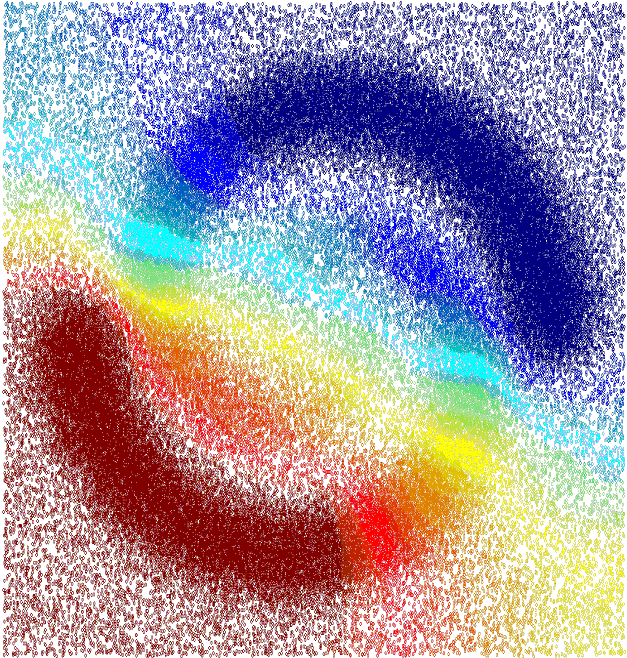}
  \label{Exp21}
  }
  \hfill
\subfigure[Output for $\beta = 0.28$.]{
  \includegraphics[width=0.3\linewidth]{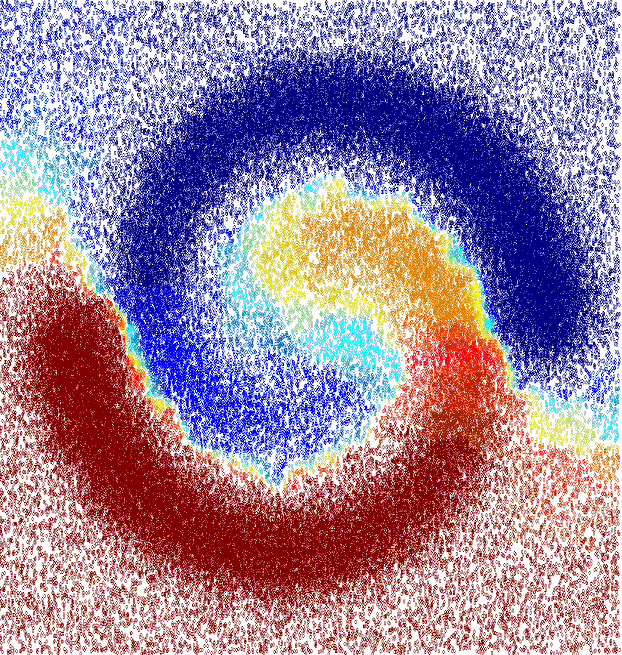}
  \label{Exp23}
}
  \hfill
\subfigure[Fuzzy Spectral Clustering.]{
  \includegraphics[width=0.3\linewidth]{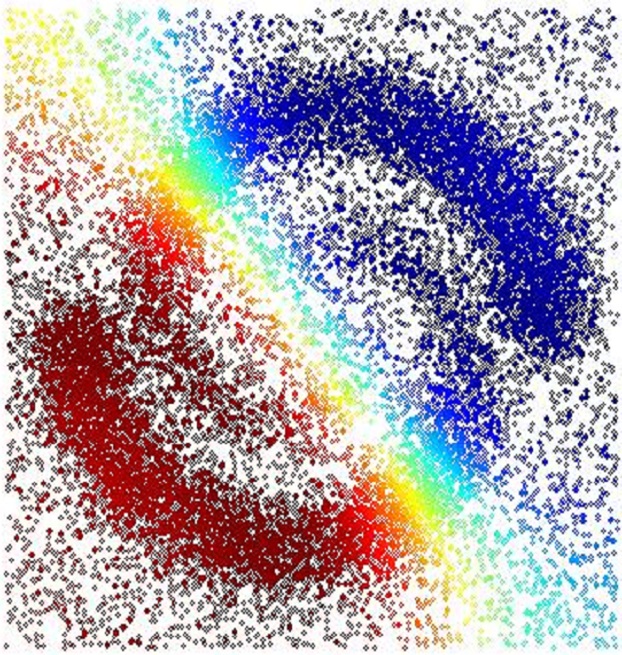}
  \label{Exp2spect}
}

\caption{Experiments on a cluttered spirals dataset.}
\label{fig:Exp2}
\end{figure}

The second dataset we consider is composed of two interleaved
spirals---see Figure~\ref{fig:Exp2}. An interesting property of this
dataset is that the head of each spiral is close (in Euclidean
distance) to the tail of the other spiral. Thus, the two clusters are
well-separated by a density gap but not by the Euclidean metric. We
use our algorithm with two different values of $\beta$: $\beta = 1$
and $\beta = 0.3$. We also run the spectral fuzzy-C means on a
subsampling of this dataset. The first thing we want to emphasize is
that the result of spectral clustering and our algorithm using $\beta
= 1$ are similar, this is to be expected as both algorithms rely on properties of
the same diffusion operator, this also means that other
fuzzy clustering techniques based on spectral clustering will fail on
this dataset.  Moreover, we can see that for $\beta = 1$, the density
gap between the two spirals is not strong enough to compensate for the
proximity of the two clusters in the Euclidean metric. On the other
hand, for $\beta \simeq 0.3$ we recover the two clusters as we give more
weight to the density structure.

\subsection{UCI datasets}

In order to obtain quantitative results regarding our fuzzy clustering scheme, we evaluate it in a classification scenario on a few datasets from the UCI repository: the Pendigits dataset ($10,000$ points and $10$ clusters), the Waveset dataset 
($5000$ points and $3$ clusters) and the Statlog dataset ($6,435$ points for $7$ clusters).  
We preprocess each dataset by renormalizing the various coordinates so they have unit variance. Then, for each dataset, we run our algorithm with various values of the parameter $\beta$ between $0.3$ and $5$, but a single value of $k$ and $\pt$ (given by a prominence gap), 
along with the fuzzy C-means algorithm for fuzziness parameters between $1.2$ and $5$. We also consider the fuzzy clustering algorithm proposed by \citet{YCEJS}, for which the cluster cores are reduced to a single point. Let $X_1,\dots,X_n$ denote our sample points and $Y_1,\dots,Y_n$ their respective labels taking 
values in $\{1,\dots,K'\}$. In these datasets, there are only two plateaus, thus we choose $\beta$.
Thus, we propose an automatic selection of $\beta$ by computing the values of the clustering entropy $H$ for multiple values of $\beta$ and by selecting 
\[
\beta = arg\,max \frac{dH}{d\beta},
\]
in other words we take $\beta$ inbetween the two plateaus by choosing the value of $\beta$ maximizing the slope of $H$. 
In order to evaluate hard clustering algorithms, it is common to use the purity measure defined by 
\[
P = \max_{\pi} \frac{1}{n}  \sum_{i=1}^n \sum_{j=1}^K 1_{\tilde{\mu}_j(X_i) = 1} 1_{Y_i = \pi(j)},
\] 
where $\pi$ is a map from the set of clusters $\{1,\dots,K\}$ to the set of labels $\{1,\dots,K'\}$.
As this measure is not adapted to fuzzy clustering, we define the \emph{$\epsilon$-entropic purity} as 
\[
HP_\epsilon = \max_\pi \frac{1}{n} \sum_i \log\left(\epsilon + \sum_{j, \pi(j) = Y_i} \tilde{\mu}_j(X_i)\right),
\]
for some $\epsilon > 0$. The $\epsilon$ parameter is used to prevent the quantity from exploding due to possible outliers.
This extension of the traditional purity can be useful to evaluate fuzzy clustering as it can be seen as an approximation of $\EE[\log(\epsilon+\sum_{j, \pi(j) = Y} \tilde{\mu}_j(X))]$ which enjoys the following property. 
\begin{proposition}
Suppose that $Y \in \{1,\dots,K\}$ and let $\epsilon > 0$, then 
\begin{multline*}
arg\,max_{f \in \RR^d \rightarrow \RR^K, \|f\|_1 = 1} \EE[\log(\epsilon+f(X))] = \\
 (1-\epsilon)^{-1} (\PP(Y = j \mid X))_{1 \leq j \leq K} - \epsilon.
\end{multline*}
\end{proposition}
Thus, for small values of $\epsilon$, a fuzzy clustering minimizing the $\epsilon$-entropic purity recovers the conditional probabilities of the labels with respect to the coordinates.

We provide the best $10^{-3}$-entropic purity obtained by each algorithm on all datasets in Table~\ref{tab:res}. As we can see, our algorithm outperforms the other fuzzy clustering algorithms on these datasets. 
In particular we can see that the simple fuzzy mode-seeking algorithm of \citet{YCEJS} fails on the Waveform dataset. 

\begin{table}
\caption{Entropic purity obtained by fuzzy clustering algorithms on UCI datasets.}
\centering
\begin{tabular}{l||c|c|c|}
\hline
 Algorithm / Data  & Waveform & Pendigits & Statlog \\
 \hline
  Ours, optimal $\beta$ & -1.1 & -0.61 & -0.51 \\
  Ours, automatic $\beta$ & -1.1 & -0.64 & -0.55 \\
  Fuzzy C-means & -1.1 & -1.35 & -0.57\\
  \citet{YCEJS} & -3.2 & -0.76 & -0.58
\end{tabular}
\label{tab:res} 
\end{table}

\paragraph{Alanine dipeptide conformations.}

\begin{figure*}[htb]
\centering
\subfigure{
  \includegraphics[width=0.30\linewidth]{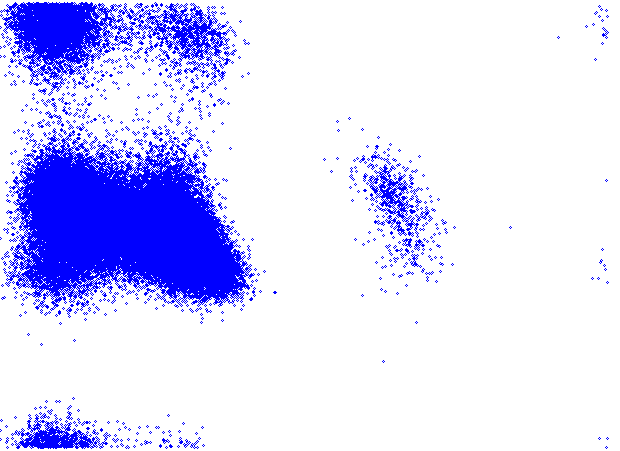}
}
\hfill
\subfigure{
  \includegraphics[width=0.30\linewidth]{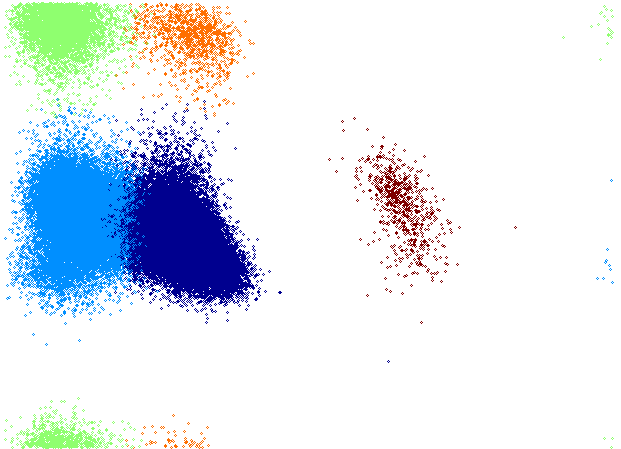}
}
\hfill
\subfigure{
  \includegraphics[width=0.30\linewidth]{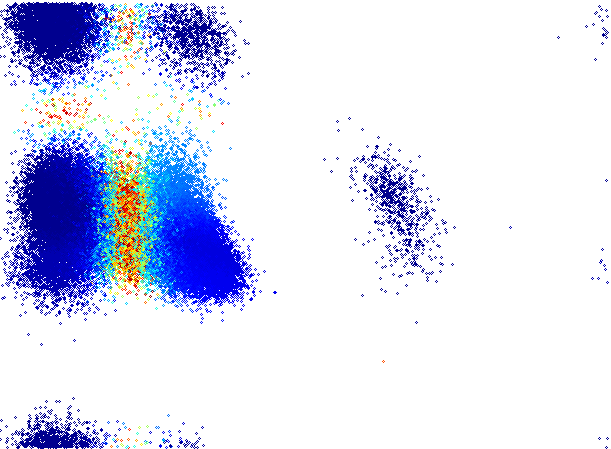}
}
\caption{From left to right: (a) the dataset projected on the Ramachadran plot, (b) ToMATo output, (c) second highest membership obtained with our algorithm for $\beta = 0.2$}
\label{fig:alanine}
\end{figure*}

We now turn to the problem of clustering protein conformations. We
consider the case of the alanine-dipeptide molecule. Our dataset is
composed of $1,420,738$ protein conformations, each one represented as
a $30$-dimensional vector. The metric used on this type of data is the
root-mean-squared deviation (RMSD). The goal of fuzzy clustering in
this case is twofold: first, to find the right number of clusters
corresponding to metastable states of the molecule; second, to find
the conformations lying at the border between different clusters, as
these represent the transition phases between metasable states.  It is
well-known that the conformations of alanine-dipeptide only have two
relevant degrees of freedom, so it is possible to project the data
down to two dimensions (called a Ramachadran plot) to have a
comfortable view of the clustering output. See
Figure~\ref{fig:alanine} for an illustration, and note that the
clustering is performed in the original space. In order to highlight
interfaces between clusters, we only display the second highest
membership function.  As we can see there are $5$ clusters and $6$ to $7$
interfaces.


\section{Proofs}
\label{sec:proofs}
\subsection{Background on diffusion processes}

Convergence of Markov chains to diffusion processes occurs in
the \textit{Skorokhod space} $D([0,T], \mathbb{R}^d)$, composed of the trajectories
$[0,T] \rightarrow \RR^d$ that are right-continuous and have left
limits, for some fixed $T>0$. 
It is equipped with
the following metric: 
\[
d(f,g) = \inf_\epsilon \{ \exists \lambda \in \Lambda, \| \lambda \| \leq \epsilon, \sup_t | f(t) - g(\lambda(t))| \leq \epsilon \},
\]
where $\Lambda$ denotes the space of strictly increasing automorphisms
on the unit segment $[0,1]$, and where
$\|\lambda\|$ is the quantity:
\[ 
\| \lambda \| = \sup_{s \neq t} \left|\log\left(\frac{\lambda(t) - \lambda(s)}{t - s}\right)\right|.
\]

In diffusion approximation, standard results prove the weak convergence of a Markov chain to a difussion process in $D([0,T],\RR^d)$. A stochastic process $M^s$ converges weakly to a diffusion
process $Y$ in $D([0,T],\RR^d)$ as $s$ tends to $0$ if and only if
\beq\label{eq:Portmanteau}
\lim_{s \rightarrow 0} \mathbb{P}(M^s \in B) = \mathbb{P}(Y \in B)
\eeq
for any Borel set $B$ such that $\mathbb{P}(Y \in \partial B) = 0$.

Let us state the convergence result when $Y$ is the Solution of the Stochastic Differential Equation~\ref{eq:diff-proc}. For this case, $b = \frac{1}{\beta} \nabla \log f$ and $a = I_d$. Consider a family of Markov chains $(M^{x_0,s})$
defined on discrete state spaces $S_s \subset \Omega$, transition kernels $K^s$
and initial states $M^{x_0,s}_0\in S_s$.  For $x \in S_s$
and $\gamma > 0$, let
\begin{itemize}
\item $a^s (x) = \frac{1}{s} \sum_{y \in S_s} K^s(x,y) (y - x) (y - x)^T;$
\item $b^s (x) = \frac{1}{s} \sum_{y \in S_s} K^s(x,y) (y - x);$
\item $\Delta_s^\gamma = \frac{1}{s} K^s(x, \mathcal{B}(x, \gamma)^c),$
\end{itemize}
where $\mathcal{B}(x, \gamma)^c$ is the complementary of the ball of radius $\gamma$ centered at $x$.

\begin{proposition}[Adapted from Theorem 7.1 in \cite{Durrett}]
\label{Durrettcor}
Let $U$ be a compact subset of $\Omega$. Let also $B$ be a Borel set in
$D([0,T], \RR^d)$ for some $T>0$ such that $\mathbb{P}(Y^{x_0} \in
\partial B) = 0$ for all $x_0\in U$. For any $\epsilon > 0$,
there exist parameters $\nu$ and $\gamma$ such that
\[
\sup_{x_0 \in U}|\mathbb{P}(M^{x_0,s}_{\lfloor t/s\rfloor} \in B) - \mathbb{P}(Y^{x_0}_{t} \in B)|
\leq \epsilon
\]
whenever the following conditions are met:
\begin{itemize}
\item[\rm (i)] $\sup_{x \in S_s} \|a^s - a\|_\infty \leq \nu;$
\item[\rm (ii)] $ \sup_{x \in S_s} \|b^s - b\|_\infty \leq \nu;$
\item[\rm (iii)] $ \sup_{x \in S_s} \Delta_s^\gamma \leq \nu;$
\item[\rm (iv)] $ \sup_{x_0\in S_s} \|M^{x_0,s}_0 - x_0\|_\infty \leq \nu.$
\end{itemize}
\end{proposition}

\subsection{Weak-Convergence}
\label{sec:weak-cvg}

In this section, we prove the following result.
\begin{proposition}
\label{weak-convergence} 
Let $Y$ be the diffusion process solution of the SDE~\ref{eq:diff-proc}. 
Let $h : \mathbb{N} \rightarrow \RR^+$ be a decreasing function such that $\lim\limits_{n \to \infty} h(n) = 0 $ and $\lim\limits_{n \to \infty} \frac{h(n)^{d+2} n}{\log n} = \infty $. 
Suppose our estimator $\hat{f}_n$ satisfies, for any compact set $U \subset \Omega$ and any $\epsilon > 0$,
\[ 
\lim\limits_{n \to \infty} \mathbb{P} (\sup_{x \in C} |f(x) - \hat{f}_n (x) | \geq h(n)^2 \epsilon) = 0.
\]
Then, for any $T,\epsilon > 0$, for any compact set $U\subset\Omega$,
and for any Borel set $B$ of $D([0,T], \RR^d)$ such that
$\mathbb{P}(Y^{y} \in \partial B) = 0$ for all $y \in U$,
there exists  a constant $C$ depending on $d$ such that for $s(n) = C h^2$, we have  
\[
\lim\limits_{n \to \infty} \mathbb{P} (\sup_{x \in U} |\mathbb{P}(M^{x,h(n)}_{\lfloor t/s(n)\rfloor} \in B) - \mathbb{P}(Y^x_t \in B)|
\geq \epsilon) = 0.
\]
\end{proposition}
The proof relies on Theorem 3 of Ting \emph{et al.} (2010) along with 
a proper control of boundary effects. Let $T$ and $\epsilon$ be strictly positive real numbers, throughout the course of the proof, the notation $M^{x,h(n)}$ stands for the continuous time process $M^{x,h(n)}_{\lfloor t/s(n)\rfloor}$.
We denote by $\mathcal{X}_n = (X_1,...,X_n)$ the i.i.d sampling which is also the state space of $M^{x,h(n)}$. 
For $\alpha > 0$, let $F^\alpha = \{ x \in \RR^d \mid f(x) \geq \alpha \}$ be the
$\alpha$ superlevel-set of $f$ and $B_\alpha = \{ w \in D([0,T], \RR^d) \mid \forall t, w(t) \in F^\alpha\}$ be trajectories staying in $F^\alpha$ up to time $T$.  
Since $Y$ does not explode in finite time, there exists $\alpha$ such that, for any $x \in U$, $\mathbb{P}(Y^x \in B_{\alpha}) \geq 1 - \epsilon / 4$.
To obtain a good approximation of the trajectories of $Y$ staying in $F^\alpha$ using $M^{x,h(n)}$, we only need to check assumptions (i)-(iv) of Proposition~\ref{Durrettcor} on $F^\alpha$. 
$F^\alpha$ is closed as $f$ is continuous and it is also bounded as $\lim_{\|x\|_2 \rightarrow \infty} f(x) = 0$, it is therefore compact. 
Applying Theorem 3 from Ting \emph{et al.} (2010) on the points of the compact set $F^\alpha$, we have, with probability $1$,
\begin{itemize}
\item[\rm (i)] $\lim\limits_{n \to \infty} \mathbb{P}(\sup_{y \in \mathcal{X}_n \cap F^\alpha} \|a^s - I_d\|_\infty \leq \nu) = 0,$
\item[\rm (ii)] $ \lim\limits_{n \to \infty} \mathbb{P}(\sup_{y \in \mathcal{X}_n \cap F^\alpha} \|b^s - \frac{\nabla{f}}{\beta f}\|_\infty \leq \nu) = 0,$
\item[\rm (iii)] $ \sup_{y \in \mathcal{X}_n \cap F^\alpha} \Delta_s^{h} = 0,$
\item[\rm (iv)] $  \lim\limits_{n \to \infty} \mathbb{P}(\|M^{x,h(n)}_0 - x\|_\infty \leq \nu) = 0.$
\end{itemize}
Thus, the assumptions (i)-(iv) of Proposition \ref{Durrettcor} are verified on $F^\alpha$.
 
Since $f$ is continuous, $B_\alpha$ is an open set. Therefore, there exists $n_0 > 0$ such that for any $n > n_0$, 
\[
\sup_{x \in U} \mathbb{P}(M^{x,h(n)} \in B_\alpha) \geq \sup_{x \in U} \mathbb{P}(Y^x \in B_\alpha) - \epsilon /4 \geq 1 - \epsilon / 2.
\]
Therefore, for any Borel set $B$,
\[
\sup_{x \in U} |\mathbb{P}(M^{x,h(n)} \in B) - \mathbb{P}(M^{x,h(n)} \in B \cap B_\alpha)| \leq \epsilon/2.
\]
Thus, we only need to approximate trajectories that do not leave $F^\alpha$ to obtain a good approximation of $\mathbb{P}(M^{x,h(n)} \in B)$. So we can apply Corollary~\ref{Durrettcor} on these trajectories with an accuracy of $\epsilon/2$ to obtain,
\[
\sup_{x \in U} |\mathbb{P}(M^{x,h(n)} \in B) - \mathbb{P}(Y^{x} \in B)| \leq \epsilon.
\]
Every step of the proof hold almost surely as $n$ tends to infinity, thus the proof of Proposition~\ref{weak-convergence} is complete.

\subsection{Proof of Theorem 1}
\label{sec:proof-fuzzycor}
Let $\beta$ and $\tau$ be strictly positive real numbers and let $U\subset \Omega$ be a compact set. 
Let $\mathcal{C}_1,\dots,\mathcal{C}_K$ be the cluster cores used by the algorithm and computed with the density estimator $\hat{f}$. 
These cluster cores are approximations of the sets $\tilde{\mathcal{C}}_1,\dots,\tilde{\mathcal{C}}_K$ obtained using the same computation with the true density $f$.
Since, by assumptions, $f$ is $\mathcal{C}^1$-continuous on $\Omega$ and $\|\nabla f\|$ is non-zero on the boundary of the $\tilde \cc$, we have 
 \begin{itemize}
\item[$(i)$] The $\tilde{\cc}_i$ are compact sets of $\RR^d$ that are well-separated (i.e. $\tilde{\cc}_{j} \cap \tilde{\cc}_{j} =
\emptyset$ for all $i \neq j$). 
\item[$(ii)$] For each $i$, the boundary of $\tilde{\cc}_i$ is smooth.
\end{itemize}
By our assumptions on the convergence of $\hat{f}$ along with Theorem 10.1 of \cite{Tomato},
\beq
\label{eq:cores-estimators}
\forall \delta > 0, \lim_{n\to\infty}\mathbb{P}(\tilde{\cc}^{-\delta}_i \subset
\cc_{i} \subset \tilde{\cc}^{\delta}_i ) = 1, 
\eeq 
where $\tilde{\cc}^{\delta}_i =
\cup_{x \in \tilde{\cc}_i} \mathcal{B}(x,\delta)$ and $\tilde{\cc}^{-\delta}_i = \tilde{\cc}^i
\setminus \cup_{x \notin \tilde{\cc}_i} \mathcal{B}(x,\delta)$.  

Without loss of generality, we can assume that $\Omega$ has a single connected component.
Let $\epsilon$ be a strictly positive real and consider $x \in \Omega$, we let
\begin{itemize}
\item $\mu^+_{i,\delta}(x)$ be the probability that $Y^x$ hits $\tilde \cc_i^\delta$ before any other $\tilde \cc_j^{-\delta}$,
\item $\mu^-_{i,\delta}(x)$ be the probability that $Y^x$ hits $\tilde \cc_i^{-\delta}$ before any other $\tilde \cc_j^{\delta}$.
\end{itemize}
Let us show that, for any $i$, a trajectory entering $\tilde \cc^\delta_{i}$ has a high 
probability to enter $\tilde \cc_i$ if $\delta$ is small enough. 
Since the $\tilde{\cc}_{i}$ are closed and disjoint there exists $\delta_0 > 0$ such that the $\tilde \cc^{\delta_0}_{i}$ are disjoints.
Moreover, since the $\tilde{\cc}_{i}$ have smooth boundaries, there exists $\delta^+_i > 0$ such that if $d(x, \tilde \cc_i) \leq \delta^+_i$ then,
the probability for $Y^x$ to hit $\tilde C_i$ before exiting $\tilde \cc^{\delta_0}_i$
is at least $1 - \epsilon /8$.

Similarly,  if a trajectory enters $\tilde \cc_i$, then it enters $\tilde \cc^{-\delta}_i$ with high probability. More precisely there exists $\delta^-_i$ such that 
if a trajectory hits $\tilde \cc_i$, then it hits $\tilde \cc_i^{- \delta}$ with probability at least $1 - \epsilon/8$.

Let $\delta = \min(\delta_j^+, \delta_j^-)$, by combining our results and using the strong Markov property of $Y^x$ we obtain that:
\begin{itemize}
\item $\mu^+_{i,\delta}(x)  - \mu_i(x) \leq \epsilon/4$,
\item $\mu_i(x) - \mu^-_{i,\delta}(x)\leq \epsilon/4$.
\end{itemize}

The next step is to show that the approximation of $\mu^+_{i,\delta}$  provided by the Markov chain is correct. 
For $T>0$, let  
\begin{multline*}
B = \{ w \in D([0,\infty], \RR^d \mid \exists \tau \text{ such that } w(\tau) \in  \tilde \cc_i^\delta \\ 
 \text{ and } \forall t < \tau \text{ we have } w(t) \in \Omega \setminus \cup_j \tilde \cc_j^{-\delta} \},
\end{multline*}
\begin{multline*}
B_T = \{ w \in D([0,T], \RR^d \mid \exists \tau \text{ such that } w(\tau) \in \tilde \cc_i^\delta  \\
\text{ and } \forall t < \tau \text{ we have } w(t) \in \Omega \setminus \cup_j \tilde \cc_j^{-\delta} \}.
\end{multline*}
We define the stopping time
\[
\tau(Y) = \inf_t Y \in \tilde \cc_i^\delta \cup_{j \in \{1,\dots,K\}, j \neq i} \tilde \cc_j^{-\delta}. 
\]
Since $\cc_i \subset \Omega$ and $\Omega$ has a single connected component, we have that $\mathbb{P}(\tau(Y^x_t) < \infty) = 1$, in particular that means that there exists $T_0$ such that 
for any $T \geq T_0$, 
$\mathbb{P}(\tau(Y^x) \leq T) \geq 1 - \epsilon/6$. Using Proposition~\ref{weak-convergence}, we have that, almost surely
$$\mathbb{P}(\mathbb{P}(\tau(M^{x,h(n)}) \leq T) \geq \mathbb{P}(\tau(Y^x) \leq T)) - \epsilon/6 \geq 1 - \frac{1}{3} \epsilon$$
Hence, we have
\begin{multline*}
\mathbb{P}(M^{x,h(n)} \in B \setminus B_T) + \mathbb{P}(Y^{x} \in B \setminus B_T) \\
\leq \mathbb{P}(\tau(M^{x,h(n)}) > T)
+ \mathbb{P}(\tau(Y^{x}) > T) \leq \epsilon/2
\end{multline*}

Since $\mathbb{P}(Y^x \in \partial B_T) = \mathbb{P}(Y^x \in \partial B) = 0$,
we can apply Proposition~\ref{weak-convergence} on the set $B_T$, and obtain 
$$\|\mathbb{P}(M^{x,h(n)} \in B_T) - \mathbb{P}(Y^{x} \in B_T)\|_{\infty, U} \leq \epsilon/4$$
Combined with our previous result, we obtain:
$$\|\mathbb{P}(M^{x,h(n)} \in B) - \mathbb{P}(Y^{x} \in B)\|_{\infty, U} \leq 3 \epsilon/4$$

Using our assumption on $\hat{C}_{i,n}$, we have $\mathbb{P}(M^{x,h(n)} \in B) \geq \hat{\mu}_{i,h,n}$. Therefore, using our previous 
bound between $\mu$ and $\mu^+$:
\[
\hat{\mu}_{i,h,n}(x) - \mu_i(x) \leq \epsilon.
\]
Similarly, 
\[
\mu_i(x) - \hat{\mu}_{i,h,n}(x) \leq \epsilon,
\]
concluding the proof. 

\section{Conclusion}

We have provided a fuzzy clustering algorithm based on the mode-seeking framework relying on the approximation of a diffusion process through the use of a random walk.
Despite the convergence issues of random-walk-based quantities for large data highlighted by \citet{gettinglost}, we have shown that our algorithm does converge to meaningful values. 
Our thereotical result is backed up by encouraging experiments. The main question still open regarding our algorithm is the choice of the temperature parameter $\beta$, while we have shown that 
the evolution of a quantification of the fuzziness of the clustering through the clustering entropy can give some hint about a correct choice for this parameter, it is not clear whether this can be done in all cases 
and for more complicated datasets.  

\paragraph*{Acknowledgements.}

The authors wish to thank Cecilia Clementi and her student Wenwei
Zheng for providing the alanine-dipeptide conformation data used in
Figure~\ref{fig:alanine}. This work was supported by the French D\'el\'egation G\'en\'erale de l'Armement (DGA), by ANR project TopData ANR-13-BS01-0008 and by ERC grant Gudhi (ERC-2013-ADG-339025).

\bibliographystyle{plainnat}
\bibliography{Density-Based_Soft_Clustering}


\end{document}